\documentclass[conference]{IEEEtran}
\IEEEoverridecommandlockouts

\usepackage{cite}
\usepackage{amsmath,amssymb,amsfonts}
\usepackage{algorithmic}
\usepackage{graphicx}
\usepackage{textcomp}
\usepackage{xcolor}
\usepackage{booktabs}

\def\BibTeX{{\rm B\kern-.05em{\sc i\kern-.025em b}\kern-.08em
    T\kern-.1667em\lower.7ex\hbox{E}\kern-.125emX}}
    
\begin{document}

\makeatletter
\def\ps@IEEEtitlepagestyle{%
  \def\@oddfoot{\mycopyrightnotice}%
  \def\@evenfoot{}%
}
\def\mycopyrightnotice{%
  {\footnotesize \begin{tabular}{@{}l@{}}
  Accepted for publication at IEEE World AI IoT Congress (AIIoT) 2026. \\
  \end{tabular}}
}
\makeatother

\title{CADENCE: Context-Adaptive Depth Estimation for Navigation and Computational Efficiency \\
}


\author{
    Timothy K Johnsen$^{1}$ and Marco Levorato$^{2}$
\thanks{
    $^{1}$Timothy Johnsen is a PhD candidate in the Computational Science joint program with San Diego State University and The University of California, Irvine in CA, USA.
        {\tt\small tjohnsen@uci.edu}
}
\thanks{
    $^{2}$Marco Levorato is a full professor in the Computer Science department at the University of California, Irvine in CA, USA.
        {\tt\small levorato@uci.edu}
}
}

\newcommand{\marco}[1]{\textbf{\color{red}[XX Marco: #1 XX]}}

\newcommand{\tim}[1]{\textbf{\color{blue}[XX Tim: #1 XX]}}

\maketitle

\begin{abstract}
Autonomous vehicles deployed in remote environments typically rely on embedded processors, compact batteries, and lightweight sensors. These hardware limitations conflict with the need to derive robust representations of the environment, which often requires executing computationally intensive deep neural networks for perception. To address this challenge, we present CADENCE, an adaptive system that dynamically scales the computational complexity of a slimmable monocular depth estimation network in response to navigation needs and environmental context. By closing the loop between perception fidelity and actuation requirements, CADENCE ensures high-precision computing is only used when mission-critical. We conduct evaluations on our released open-source testbed that integrates Microsoft AirSim with an NVIDIA Jetson Orin Nano. As compared to a state-of-the-art static approach, CADENCE decreases sensor acquisitions, power consumption, and inference latency by 9.67\%, 16.1\%, and 74.8\%, respectively. The results demonstrate an overall reduction in energy expenditure by 75.0\%, along with an increase in navigation accuracy by 7.43\%.
\end{abstract}


\section{Introduction}


Small autonomous vehicles, like drones, rely on lightweight sensors that often lack the precise 3D environmental data required to navigate unknown terrains. To compensate, modern systems use deep neural networks (DNNs) for complex computer vision tasks like monocular depth estimation (MDE) and object detection. However, the high computational demands of these DNNs clash with limited onboard resources, risking mission failure through battery depletion or delayed processing.

Current research in mitigating such computational demands follows two primary methodologies. The first, static model reduction, simplifies DNNs via pruning \cite{blalock2020state}, knowledge distillation \cite{hinton2015distilling}, quantization \cite{pappalardo2022qonnx}, and direct design \cite{gholami2018squeezenext}. While effective, they often incur a degradation in task accuracy. Further, they must be complex enough to handle the most challenging scenarios, which leads to suboptimal efficiency in terms of inference latency, power consumption, and energy expenditure when encountering simpler scenarios. The second approach, edge computing \cite{VideoEdge}, offloads perception tasks to remote servers, which introduces critical vulnerabilities when establishing and maintaining volatile communication links in terms of latency, data privacy, and connectivity. 

To address these challenges, we adopt a third methodology that adapts dynamic DNNs \cite{johnsen2024overview} to scale computations at runtime in response to context. This removes the need for volatile communication links as used in edge computing, and mitigates the suboptimal efficiency and accuracy degradation inherent in static model reduction. The primary challenge of this framework is in developing the adaptive logic.

\begin{figure}[h]
\centerline{\includegraphics[width=0.45\textwidth]{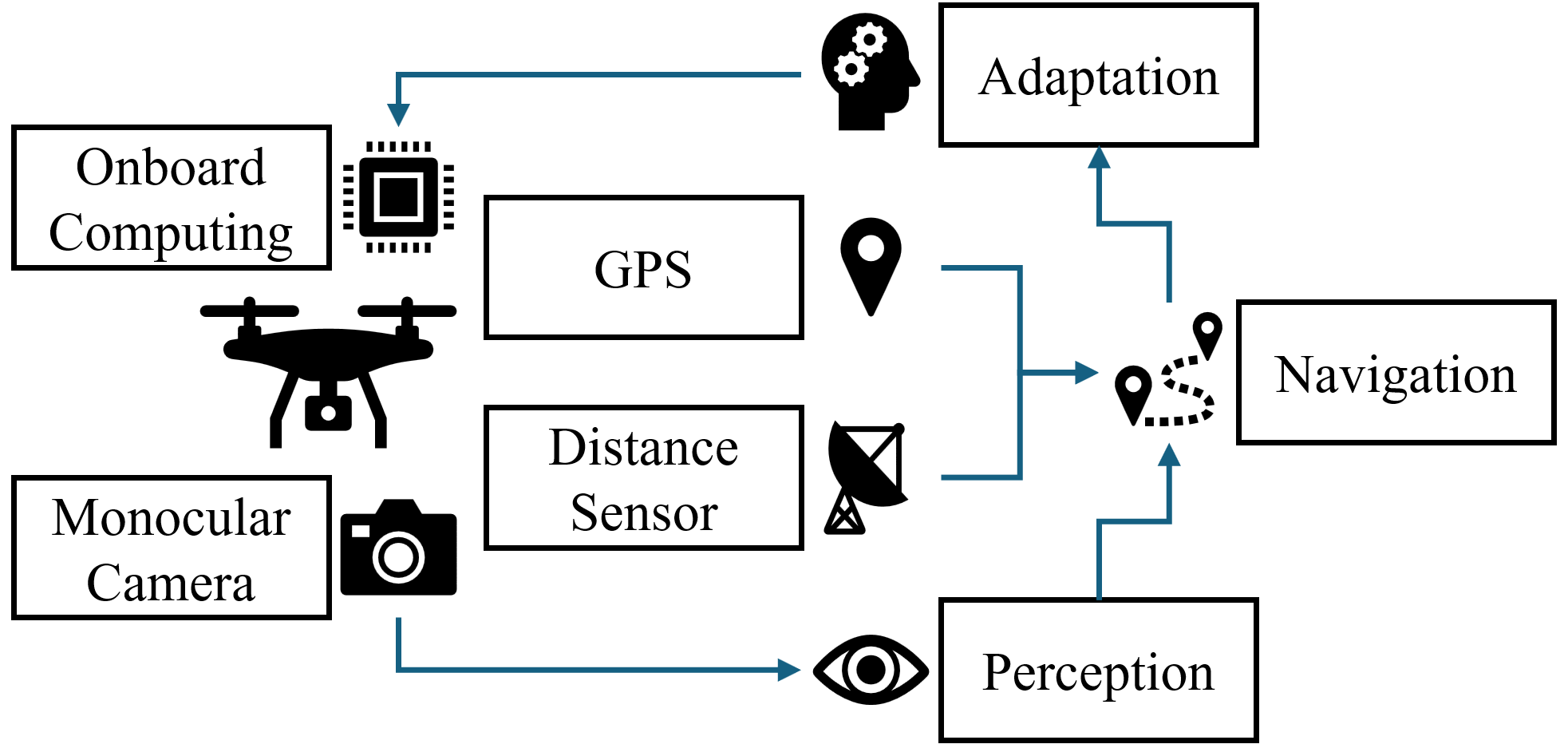}}
\caption{System environment that contains an autonomous drone equipped with an embedded computer, GPS, distance sensor, and monocular camera.}
\label{fig:environment}
\end{figure}

Our objective is to optimally navigate a drone through an unknown terrain while minimizing computational overhead (as shown in Fig.~\ref{fig:environment}). Specifically, we present CADENCE (Context-Adaptive Depth Estimation for Navigation and Computational Efficiency), a full autonomy stack that closes the sensing-actuation loop by dynamically scaling perception fidelity in response to navigation and environmental context.

\subsection{Contributions}

\noindent
$\bullet$ We modify a state-of-the-art (SoTA) MDE architecture \cite{hu2021single} by training it to be a slimmable network, allowing for online dynamic scaling of computational resources at runtime.

\noindent
$\bullet$ We propose a unified navigation-and-adaptation policy that jointly predicts motion actions and slimming factors, eliminating the computational overhead of two separate auxiliary and navigation policies as adopted in SoTA models \cite{johnsen2024navislim}.

\noindent
$\bullet$ We present CADENCE, a full autonomy stack that adapts computing efficiency in response to case-by-case context. We further develop a robust multi-stage training procedure.

\noindent
$\bullet$ We release a hardware-in-the-loop (HIL) Python repository\footnote{https://github.com/WreckItTim/OmniNaviPy}. that integrates Microsoft AirSim \cite{shah2018airsim}, NVIDIA Jetson benchmarking code, MDE with slimmable networks, collected datasets, trained models, and hold-out evaluation tools.

As compared to a non-adaptive SoTA pipeline, CADENCE decreases the number of sensor acquisitions by 9.67\%; decreases the inference latency, power consumption, and energy expenditure of the MDE network by 74.8\%, 16.1\%, and 75.0\%; and increases the navigation accuracy by 7.43\%.

\section{Related Works}

SoTA approaches enable autonomous robotics via deep reinforcement learning (DRL) with RGB-Depth inputs \cite{vemprala2021representation}. However, such applications typically rely on static DNNs, requiring the model complexity to match the most challenging operating conditions. This results in unnecessarily high resource usage during less demanding situations. In contrast, we present a dynamic approach where a portion of the computing is scaled at runtime, maintaining high-fidelity perception only when necessary for navigation reliability. 

Dynamic DNNs \cite{johnsen2024overview} scale operations using early exits \cite{matsubara2022split}, slimmable networks \cite{yu2018slimmable}, or other architectural attributes. Unlike methods that swap models at runtime and subsequently incur context-switching latency, dynamic DNNs are loaded before runtime. Only a few approaches exist in literature that adapt scaling in response to intrinsic latent features or extrinsic environmental factors \cite{malawade2022hydrafusion, odema2022testudo, jiang2023dynamic}. Previous work \cite{johnsen2024navislim} explores adaptive onboard navigation using slimmable MLPs; however, MLPs offer smaller efficiency gains compared to CNNs, and \cite{johnsen2024navislim} ignores the computational overhead of perception DNNs, like MDE, by relying on ground truth depth information derived from onboard depth sensors.

To fit lightweight devices, SoTA MDE networks \cite{hu2021single} employ static model reduction or edge computing, such as done in \cite{zhang2024repmono} with direct design, \cite{xiao2024improving} with knowledge distillation, and \cite{johnsen2024navisplit} with adaptive split computing. While dynamic DNNs have been applied to MDE in isolated cases (\textit{e.g.,} via early exits \cite{cipolletta2021energy}), methodologies for selecting onboard scaling factors based on navigation goals and environmental factors remain absent in literature. To our knowledge, CADENCE is the first approach to adapt onboard dynamic MDE networks to improve computational efficiency during autonomous navigation.

\begin{figure*}[t]
\vspace{0.1in}
\centerline{\includegraphics[width=1.0\textwidth]{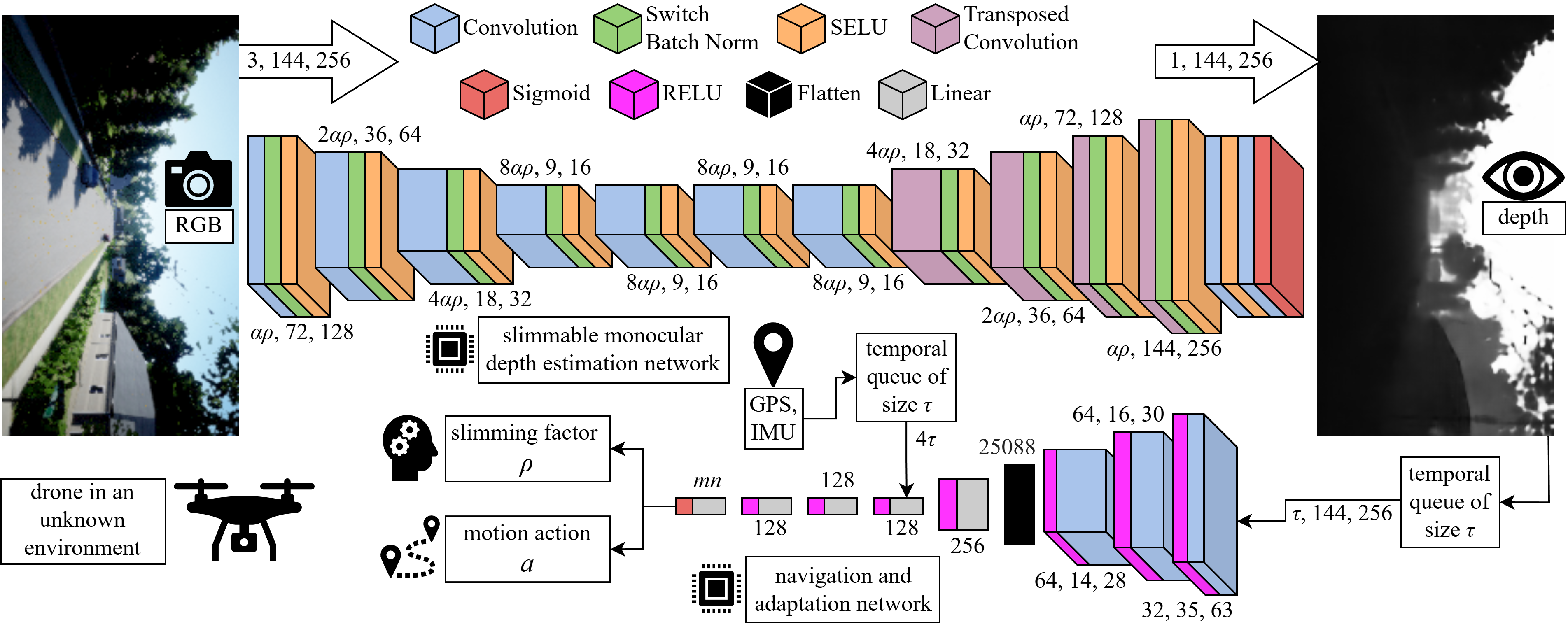}}
\caption{Illustrated is the flow of data from raw sensor acquisition to intelligent decision-making through the full autonomy stack, CADENCE.}
\label{fig:system-model}
\end{figure*}

\section{Problem Formulation}
\label{sec:problem}

Consider the development of policy $\phi$ that generates sequences of actions to both reliably and efficiently reach a target position within an unknown terrain. Executing the policy results in stepping through consecutive sensing-computing-acting stages that continue until the vehicle reaches a target (success) or a time limit is exceeded (failure).

During the sensing stage, a set of observations are acquired from onboard sensors that are added to a first-in-first-out (FIFO) queue, $X$, of temporal length $\tau$. During the computing stage, $X$ is processed via an adaptive pipeline to predict: (1) motion actions, $A$, that place the vehicle on a length-optimal trajectory towards the target; and (2) a gate value, $G$, that corresponds to a specific subset of model parameters, $\hat{\theta} \subseteq \theta$, to be utilized by the pipeline at the next computing stage. During the acting stage, the vehicle executes $A$ and allocates the proper computing resources as required by $G$.

We consider that the entire processing pipeline is split into: (1) a dynamic component that has scalable computing capabilities, and (2) a downstream static component that receives input from the dynamic portion and further processes it for final output values. The dynamic portion of the processing pipeline, $f$, produces an intermediate output, $Y$, at step $t$:
\begin{equation}
\label{eq:dynamic}
\begin{aligned}
Y^{(t)} = f_\theta(X_f^{(t)}, G^{(t)}),
\end{aligned}
\end{equation}
where $X_f^{(t)}$ is the subset of $X$ required to compute $f$, and $Y$ is input into the static portion of the processing pipeline, $g$:
\begin{equation}
\label{eq:CADENCE}
\begin{aligned}
[A^{(t)}, G^{(t+1)}] = g_\psi(Y^{(t)}, X_g^{(t)}),
\end{aligned}
\end{equation}
where $\psi$ represents the model parameters used by $g$, and $X_g^{(t)}$ is the subset of $X$ required to compute $g$. We further formulate the objective for CADENCE as an optimization problem:

\begin{equation}
\label{eq:opt-dual}
\begin{aligned}
argmin_{\phi} \; \left\langle E_\phi \right\rangle \\
\textrm{s.t.} \; \left\langle \eta_\phi \right\rangle \geq \eta_0
\end{aligned}
\end{equation}
where $\left\langle E_\phi \right\rangle$ is the expected efficiency when operating under policy $\phi$, $\eta$ denotes reliability, and $\eta_0$ is a constant representing a baseline reliability threshold. In this work, reliability $\eta$ represents navigation accuracy, with $\eta_0$ set to the maximum accuracy achieved from a static, high-fidelity baseline. Efficiency $E$ is defined as the energy expended during the computing stage, which is correlated to power consumption and inference latency. The motion actions $A$ control an autonomous drone, and the gate control $G$ corresponds to scaling a slimmable MDE network. The static portion of the computing stage is a navigation-and-adaptation policy used to transform the estimated depths, IMU, and GPS data into $A$ and $G$.

\section{Methods}
\label{sec:methods}

The CADENCE architecture is illustrated in Fig.~\ref{fig:system-model}, and consists of two primary components. The perception component consists of a slimmable MDE network that not only accurately extracts 3D depth information from a monocular RGB image, but is specifically trained so that the width of the DNN can be scaled at runtime for dynamic fidelity and computing efficiency. The logic component consists of a unified navigation-and-adaptation policy trained with DRL to generate long-horizon actions that control both the robot's trajectory and the scaling of the upstream perception component.

\subsection{Perception: Slimmable MDE Network}
\label{sec:perception}

Perception extracts a 2D depth map, $\widehat{\mathbf{D}}$, from an RGB image, $\mathbf{I}$. We use a CNN hyper-parameter structure like that used in DGNLNet~\cite{hu2021single}. Standard MDE networks are inherently static and require full execution at every inference, leading to significant energy waste and suboptimal inference latency. We transform this architecture into a slimmable network that can dynamically scale its computational overhead at runtime.


Let $\boldsymbol{\rho} = [\rho_1, \rho_i, ...., \rho_n]$ be a set of selectable slimming factors whose values are each in the range (0, 1], representing the percent of active channels in each hidden layer. This dynamic portion of the computing stage, $f$, is defined as:

\begin{equation}
\label{eq:model-depth}
\begin{aligned}
\widehat{\mathbf{D}} = f_\theta(\mathbf{I}, \rho),
\end{aligned}
\end{equation}
where $\theta$ represents the optimized model parameters. Unique to CADENCE, when $\rho=0$, the system entirely bypasses image acquisition and network execution, instead providing a zero-filled depth map. This state represents the maximum energy-saving mode, quickest inference time, and smallest power consumption. The slimmable MDE network is trained by minimizing the following loss function:

\begin{equation}
\label{eq:opt-depth}
\begin{aligned}
argmin_{\theta} \; \left\langle loss(\mathbf{D}, \widehat{\mathbf{D}}^{(\rho)}) \; \forall \; \rho \in \boldsymbol{\rho} \right\rangle,
\end{aligned}
\end{equation}
where $\widehat{\mathbf{D}}^{(\rho)}$ is the predicted depth map when using the given slimming factor $\rho$, and $\mathbf{D}$ is the ground truth depth map. There are unique challenges when optimizing $\theta$ both as a slimmable DNN, and for MDE. At each iteration of the optimizer, the gradient must now be calculated $n$ number of times (for each value of $\rho$). This can lead to unstable gradients, which is why we use switch batch normalization \cite{yu2018slimmable} to calculate the network statistics separately for each value of $\rho$. For the specific task of MDE, we further use weight decay and L1-loss, common in literature for training static MDE networks \cite{hu2021single}. Note that the maximum number of channels in each hidden layer is controlled by a constant scaling value, $\alpha$, as shown in Fig.~\ref{fig:system-model}, which is a static value set before runtime. At runtime, the dynamic value of $\rho$ is set during inference.

\subsection{Navigation and Adaptation Policy}
\label{sec:methods-navi}

We propose a unified navigation-and-adaptation policy (\ref{eq:policy}) that jointly predicts the motion action $a$ and slimming factor $\rho$. This joint formulation addresses the inherent interdependence between perception and control: $\rho$ changes the quality of $\widehat{\mathbf{D}}$ which directly impacts the prediction of optimal actions $a$. Conversely, $a$ alters the drone's trajectory and future environmental context, which may alter optimal values of $\rho$.

\begin{equation}
\label{eq:policy}
\begin{aligned}
[a, \rho] = g_\psi(\widehat{\mathbf{D}}, \textbf{p}),
\end{aligned}
\end{equation}
where $\textbf{p}$ is the relative pose of the drone w.r.t. the target.

Fig.~\ref{fig:system-model} illustrates the DNN structure for the joint navigation-and-adaptation policy, which is significantly smaller than the primary MDE DNN. The architecture features a CNN backbone that processes input depth maps. Its output latent features are flattened and concatenated with the drone's relative pose, derived from onboard GPS and IMU. To incorporate temporal context, observations are structured as a FIFO queue storing the $\tau$ most recent time steps. Inspired by \cite{mnih2015human}, the CNN backbone accepts $\tau$ input channels, and the relative poses are similarly concatenated across these $\tau$ steps. For the initial steps of a trajectory, missing observations are zero-padded (a value reserved during normalization). Aligning with existing literature \cite{mnih2015human}, we empirically find $\tau=3$ to be highly robust.

Realizing the optimal slimming factors is an intractable problem, due to the infinite state-space resulting from the inherent interdependence between $a$ and $\rho$. Thus we train the policy using DRL, specifically with a double Deep Q-Network (DQN) \cite{van2016deep}. This DNN structure maps a continuous input to discrete output. We shape a robust reward function as found through empirical findings with surrogate models:
\begin{equation}
\label{eq:reward-navi}
    reward = 
\begin{cases}
      -10 & \text{time constraint} \\
      40 & \text{goal} \\
      -d -E(\rho) - 1 & \text{otherwise}
\end{cases}
\end{equation}
This applies terminal conditions based on task success: a penalty if the episode violates the time constraint, and a reward if the drone reaches the target. For non-terminal steps, it applies penalties to encourage efficient progress: $d$ penalizes the Euclidean distance to the target; $E(\rho)$ penalizes the energy expended by the slimmable MDE network on an NVIDIA Jetson Orin Nano (constituting the HIL component of CADENCE); and a constant penalty discourages long episodes.

Each output node of the DQN corresponds to a paired motion action and slimming factor. There are $n*m$ many output nodes, where $n$ corresponds to the number of selectable values of $\rho$ and $m$ corresponds to that of $a$. The DQN selects from several discrete magnitudes in either direction, and in rotations. Translational moves with larger magnitudes, measured in meters, will correspond to longer intervals in-between steps. The drone executes the action in real-time until either the magnitude of the action is fulfilled or a rigid surface is detected via a short-range distance sensor, thus the magnitude determines a variable navigation frequency $\nu$. 

We use a curriculum learning schedule to incrementally increase the difficulty of episodes used to train the DQN. A unique episode is defined by its starting and target positions. Each time the curriculum learning schedule ``levels up", it increases the maximum difficulty that unique episodes can be sampled from. We level up after every 10,000 sampled episodes, and set a $70\%$ chance to sample an episode from the highest difficulty and a $30\%$ chance to sample from any lower difficulty. We sample from lower difficulties to avoid catastrophic forgetting, which is a phenomenon encountered in DRL where the policy will forget previously learned behavior if old data are no longer in the replay buffer. After the highest level is reached, all difficulty levels are sampled from using a uniform distribution. The learning loop terminates after two million episodes. Every 10,000 episodes, the model is evaluated against a static validation set. After training, the DQN weights corresponding to the highest validation accuracy are used for final evaluations on the hold-out test set.

We utilize an A* shortest path algorithm \cite{hart1968astar} to generate a set of ground truth length-optimal paths. The paths are sorted by difficulty, as determined by the number of actions required to reach the target. These are only used to: (a) determine if there is a viable path between randomly proposed start and target locations, (b) evaluate the difficulty of a path, and (c) generate static hold-out sets used for validation and testing.

\section{Results}
\label{sec:results}

We train and evaluate CADENCE using our released repository that integrates a drone simulator with tools for machine learning, benchmarking, data collection, processing, and visualization. Evaluations focus on: (1) quantifying the efficiency gains on embedded IoT hardware, and (2) assessing the impact of adaptive perception on navigation reliability.

\subsection{Testbed}
\label{sec:results-testbed}

To bridge the gap between simulation and real-world deployment, we utilize a HIL testbed. The navigation environment is simulated using Microsoft AirSim \cite{shah2018airsim} which provides high-fidelity physics and photo-realistic RGB-Depth sensor data. CADENCE is executed on an NVIDIA Jetson Orin Nano to evaluate computing efficiency and define the energy function in equation~\ref{eq:reward-navi}. We conduct experiments using the ``AirSimNH" map, which simulates an urban environment with houses, cars, and trees. We use the bottom-right three quadrants of the map for training and the top-left one for validation and testing.

Training a DQN to convergence is highly time-intensive due to the vast number of required episodes, simulator interfacing latency, and DRL's sensitivity to random seeds (which necessitates multiple restarts). Consequently, surrogate models (which do not train to completion on a full dataset) are needed during reward shaping, and elaborate iterative and comparative processes (which analyze fine-grained effects of various parameters and methods) are not feasible. To accelerate the training process, we distribute the workload across multiple servers and bypass simulator latency entirely using a custom data tool (provided in our released repository) that caches sensor observations based on discretized map locations.

\subsection{Benchmarking}
\label{sec:results-benchmarking}

\begin{table}[htbp]
\caption{}
\begin{center}
\begin{tabular}{cccccc}
\toprule
$\alpha$ & power [mW] & latency [ms] & energy [mJ] \\ 
\midrule
256 & 17740 & 117.1 & 2078.1 \\
128 & 18351 & 30.0 & 550.0 \\
64 & 17808 & 11.0 & 196.4 \\
32 & 12961 & 6.5 & 83.9 \\
16 & 7375 & 7.0 & 51.6 \\
8 & 6123 & 6.7 & 41.1 \\
4 & 5224 & 6.5 & 34.1 \\
2 & 4968 & 6.0 & 29.8 \\
1 & 4788 & 6.0 & 28.7 \\
\bottomrule
\end{tabular}
\label{tab:benchmarks}
\end{center}
\end{table}

The primary objective of our framework is to reduce the computational bottleneck imposed by MDE. Table~\ref{tab:benchmarks} summarizes the performance of the MDE network across its various sizes as defined by $\alpha$ (see Fig.~\ref{fig:system-model}) with a fixed value of $\rho=1$. The power consumption is directly queried using the native NVIDIA API and is measured across the entire board, including effects from static memory storage and other auxiliary computational costs. Benchmarking is conducted by iteratively processing a set of RGB images and measuring the resulting inference latency and power consumption. This process is repeated multiple times, and the order is randomized to mitigate effects of running the processes over a long period. Both the total inference and power is averaged over all forward passes. Energy is calculated by integrating power multiplied by latency. At $\alpha$ values between 64 to 256, the maximum power is being used; however, there are quite significant changes in inference latency. Alternatively, at $\alpha$ values between 4 to 32, there are more significant changes in power consumption rather than in inference latency. At $\alpha$ values below 4, there are nominal changes in both power and latency. This displays an interesting set of profiles that depend on the network size. 

To provide context, executing the MDE network at maximum power, $\alpha > 32$, consumes nearly one-third of the total energy capacity onboard a commercially available drone on today's market.  This overhead significantly reduces flight duration and increases the risk of mission failure. However, scaling down the $\alpha$ value reduces inference latency, which simultaneously lowers energy expenditure and improves reaction time. Furthermore, CADENCE's variable navigation frequency (Section~\ref{sec:methods-navi}) minimizes the total number of MDE executions required, further preserving the energy reservoir.




\subsection{Slimmable MDE Network}
\label{sec:results-depth} 

We compare several configurations of the slimmable MDE network to analyze the trade-offs between depth accuracy, energy consumption, and the navigation policy. 

\begin{figure*}[t]
\centerline{\includegraphics[width=1.0\textwidth]{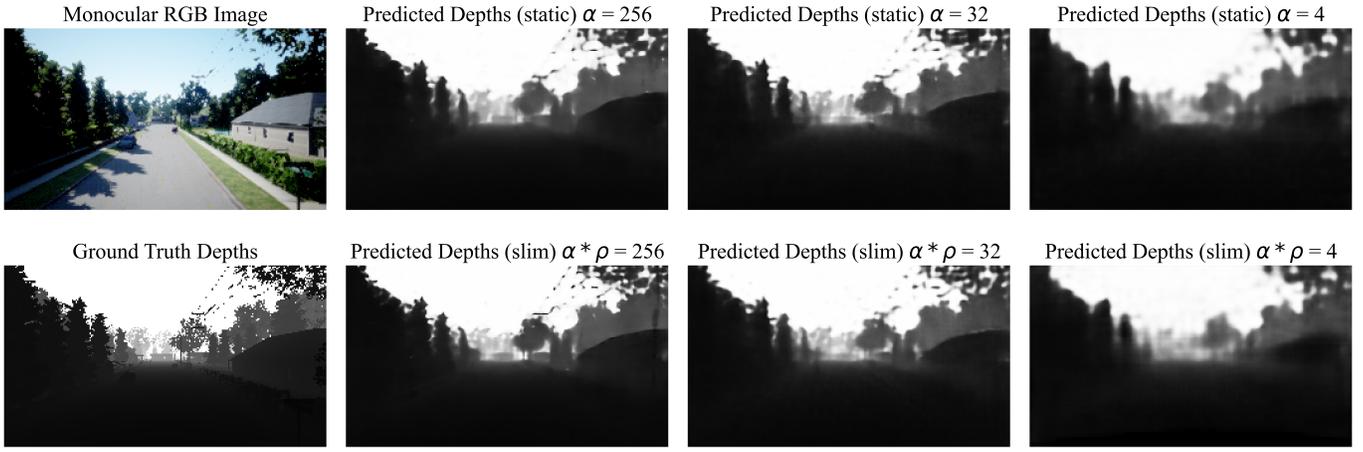}}
\caption{Example RGB image, ground truth depth map, and predicted depth maps for both static and slimmable networks with varying network sizes.}
\label{fig:depth-example}
\end{figure*}

We sample a dataset of $20,000$ RGB-Depth pairs, and split them into $40\%$ training, $10\%$ validation, and $50\%$ testing. Fig.~\ref{fig:depth-example} compares an RGB image, the ground truth depth, and depths estimated from either a static or slimmable network. Network size is calculated by multiplying $\alpha * \rho$. Depth predictions are more blurred around the edges of objects at lower network sizes -- illustrating advantages of higher fidelity MDE.

\begin{figure}[h]
\centerline{\includegraphics[width=0.45\textwidth]{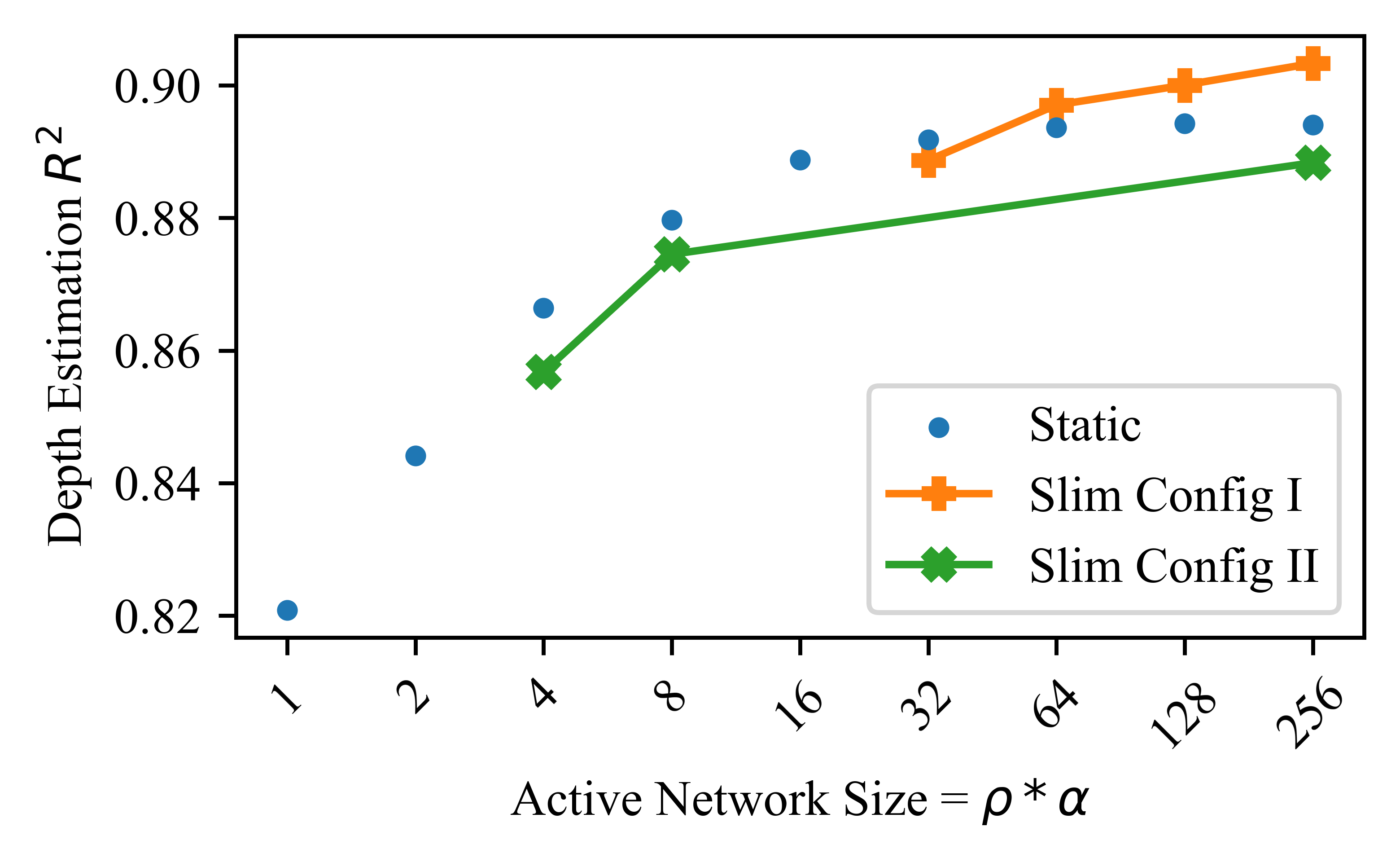}}
\caption{Test $R^2$-scores of various trained MDE network configurations.}
\label{fig:depth-curves}
\end{figure}

Results indicate a strong linear correlation between expected and predicted depth values, achieving $R^2$-scores up to 0.9033. High errors occasionally occur at large depth magnitudes, primarily along object edges and within sparse structures like leaves, causing the blurring artifacts shown in Fig.~\ref{fig:depth-example}. Fig.~\ref{fig:depth-curves} compares the $R^2$-scores of static and slimmable MDE networks across two configurations ($\alpha=256$). Configuration I utilizes the four largest sizes with slimming factors $\boldsymbol{\rho} = [1, 1/2, 1/4, 1/8]$, while Configuration II adopts an aggressive reduction strategy using $\boldsymbol{\rho} = [1, 1/32, 1/64]$. Consistent with findings in \cite{yu2018slimmable}, the slimmable networks achieve higher $R^2$-scores than static baselines at their largest sizes. Expanding the range of slimming factors (as seen in Configuration II), along with the number of slimming factors (not shown for brevity), degrades overall accuracy relative to the static networks.

\subsection{Navigation and Adaptation}
\label{sec:results-navi}

\begin{figure}[h]
\centerline{\includegraphics[width=0.5\textwidth]{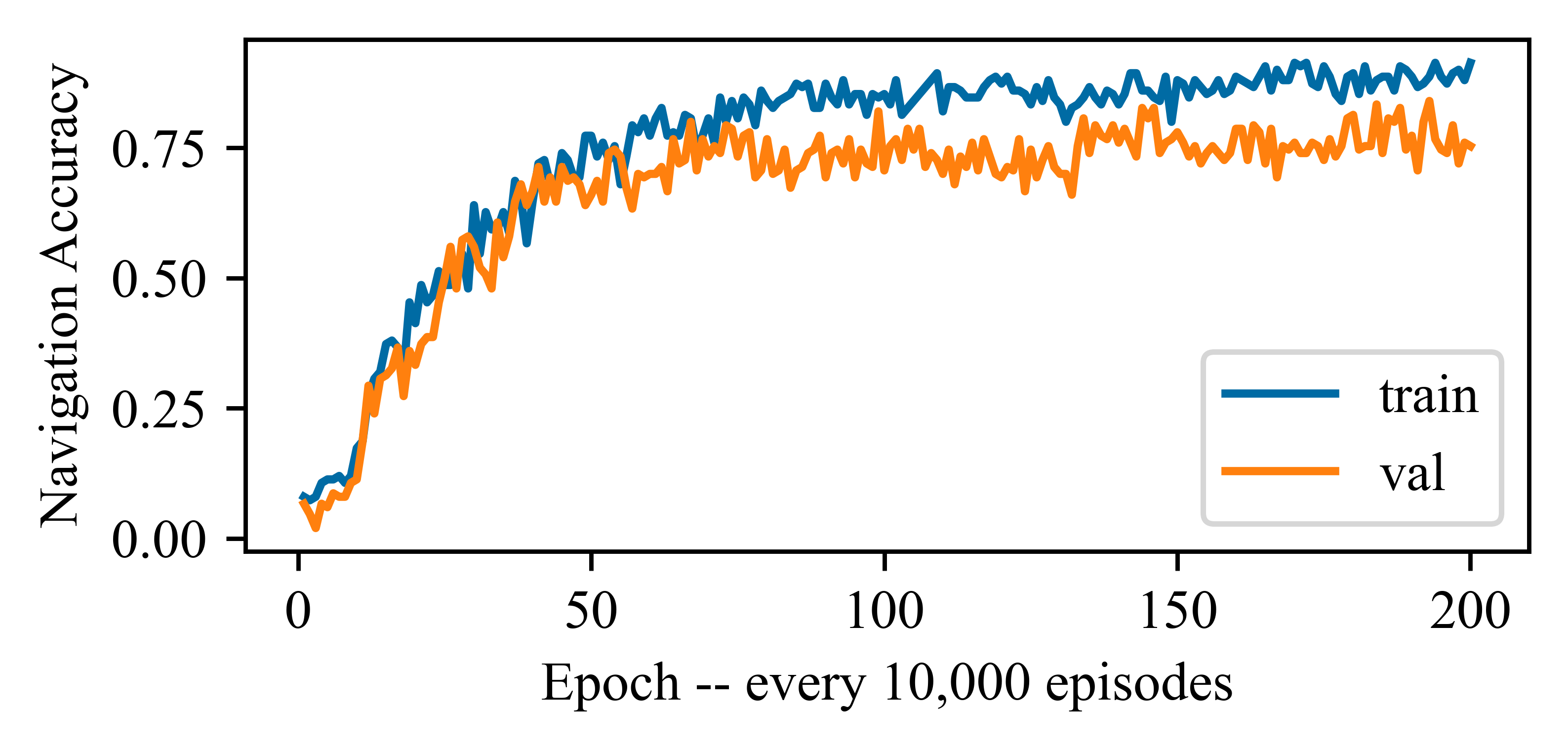}}
\caption{Learning curve from training the navigation-and-adaptation policy.}
\label{fig:navi-learning}
\end{figure}

Fig.~\ref{fig:navi-learning} shows a learning curve after training the navigation-and-adaptation policy with DRL. The validation accuracy converges as it begins to overfit to the training data. Early stopping is applied by selecting the model state corresponding to the highest validation accuracy.

\begin{figure}[h]
\centerline{\includegraphics[width=0.5\textwidth]{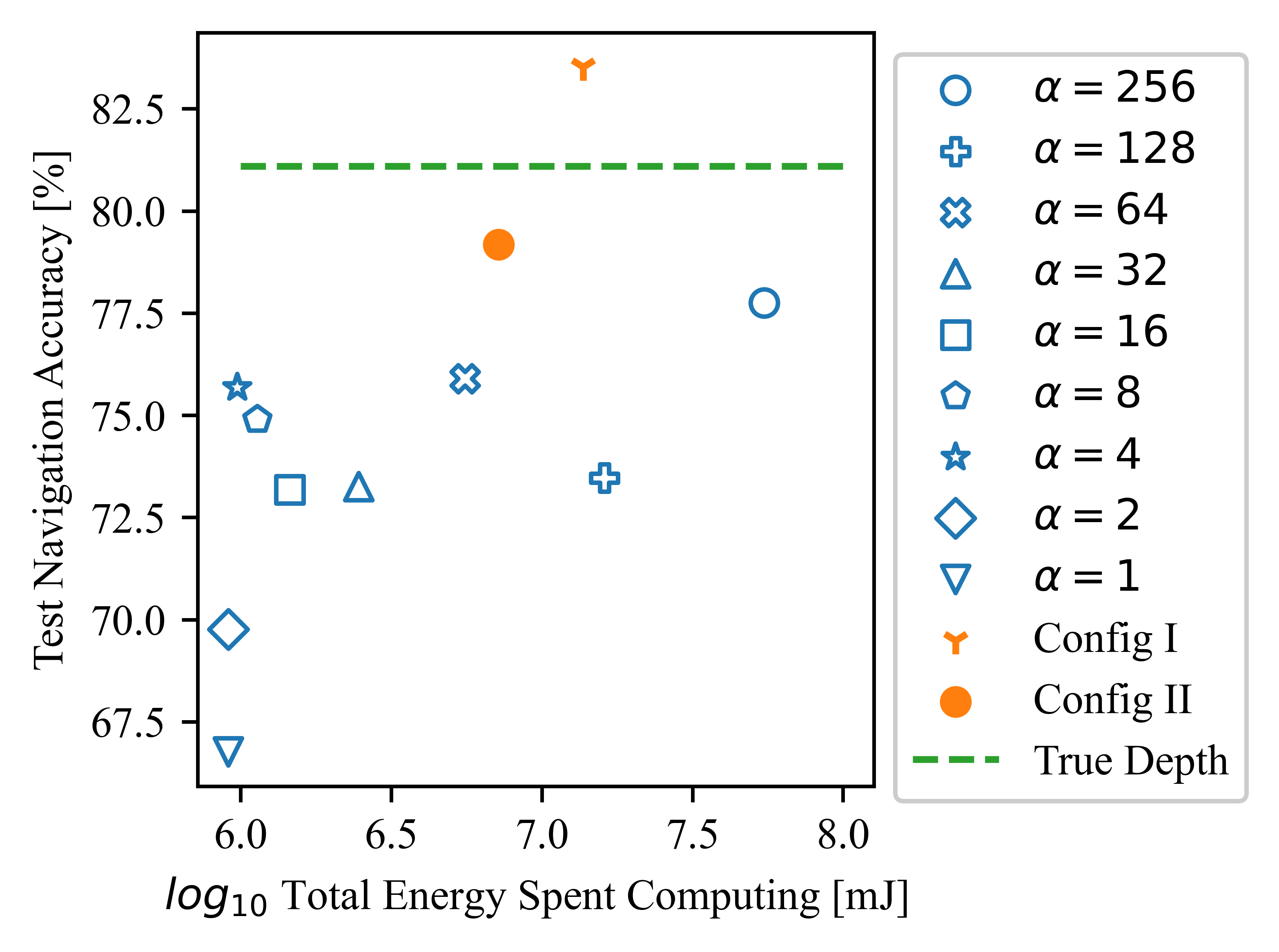}}
\caption{Final evaluations on the static holdout test set of ground truth A* paths, for each trained policy. The x-axis shows the energy expended while computing the MDE network over all paths. The hollow blue markers correspond to using a SoTA static MDE network with size $\alpha$, and a navigation DQN policy that predicts $a$. The solid orange markers correspond to using a slimmable MDE network with a navigation-and-adaptation DQN that predicts both $a$ and $\rho$. The green dashed line corresponds to a navigation DQN that bypasses MDE and instead directly uses the AirSim ground truth depth maps.}
\label{fig:energy-accuracy}
\end{figure}

Fig.~\ref{fig:energy-accuracy} highlights a key outcome of CADENCE, presenting two solutions to the optimization problem in (\ref{eq:opt-dual}). Prioritizing navigation accuracy using solely static SoTA models necessitates selecting the largest variant ($\alpha=256$), which incurs substantial energy costs. Both dynamic adaptation configurations outperform the largest static network in terms of energy efficiency and navigation accuracy. Furthermore, the accuracy decline w.r.t. smaller static networks warrants the need for complexity adaptation. Notably, Configuration I achieves higher accuracy than even the ground truth depth data. We posit that the partial blurring inherent in smaller network outputs acts similarly to Gaussian noise injection, enhancing model robustness and generalization. Consequently, blurring presents a trade-off: it risks underfitting when tight edge navigation is required, but mitigates overfitting on novel data. By providing multiple network sizes, CADENCE empowers the adaptation logic to dynamically balance these effects based on the immediate environmental context.

To validate this theory, we analyze the distinct paths successfully navigated by different static DQN policies. Let $P$ be the complete set of test paths, and $A_i \subset P$ be the subset of paths successfully completed by the $i$-th model. We define $B = \bigcup_{i=1}^n A_i$ as the union of unique successful paths across $n$ independent policies. Our goal is to saturate $B$ utilizing the fewest number of policies (i.e., where increasing $n$ yields only marginal gains). Iteratively adding models in descending order of network size achieves saturation at $\alpha=32$, requiring policies with $\alpha \in \{256, 128, 64, 32\}$. This is Configuration I (see Fig.~\ref{fig:energy-accuracy}). A greedy approach, iteratively adding the model that most significantly expands $B$, regardless of network size, saturates with just three policies: $\alpha \in \{256, 8, 4\}$, mirroring Configuration II. These saturation dynamics prove that while two policies may exhibit similar overall accuracies, their path-solving behaviors differ drastically. We posit that this behavioral variance stems from size-dependent blurring, which the adaptation policy controls across diverse scenarios.

\begin{figure}[h]
\centerline{\includegraphics[width=0.5\textwidth]{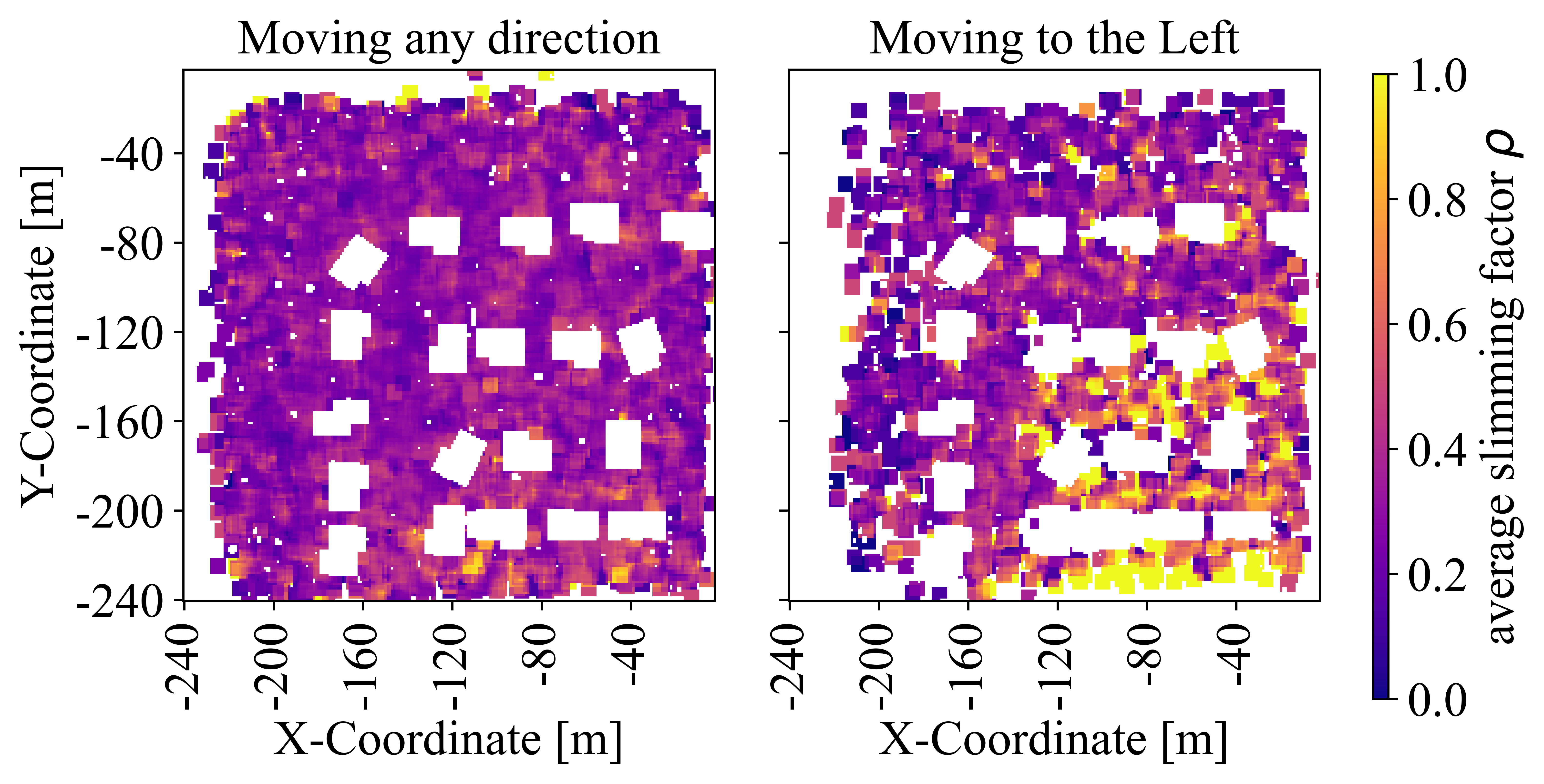}}
\caption{Average slimming factor when computing the slimmable MDE network at each of the given (x, y) locations. Compared is the drone moving in an arbitrary direction versus moving only to the left.}
\label{fig:egrid}
\end{figure}

DNNs inherently have a lack of understanding into why predictions are made. This leads to safety and reliability issues when deploying to control algorithms used by autonomous vehicles. We examine correlations that may help illuminate why the trained navigation-and-adaption policy (Configuration I) predicts particular values of $a$ and $\rho$. Fig.~\ref{fig:egrid} shows the average predicted slimming factor throughout the test region, indicating that areas with higher densities of houses correspond to higher values of $\rho$. This correlation becomes more prominent when the drone is on a trajectory moving through small alleyways in-between houses (moving to the left). This warrants the advantage of having higher fidelity depth maps to better navigate between objects, and that the trained policy is considering navigation objectives along with the current trajectory when predicting the optimal adaptation value $\rho$.

\begin{figure}[h]
\centerline{\includegraphics[width=0.48\textwidth]{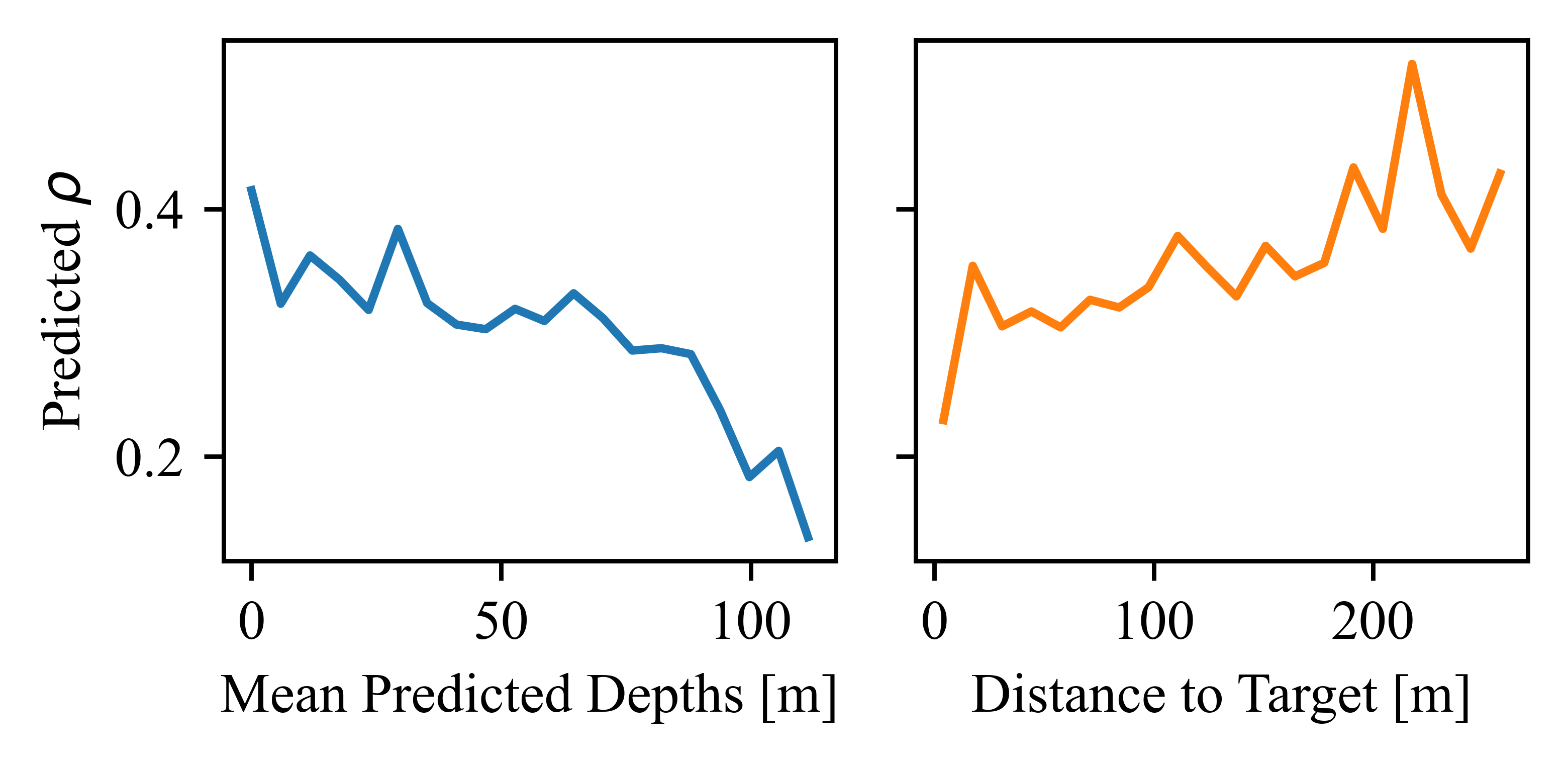}}
\caption{Correlation between the adaptation parameter $\rho$ and two environmental factors: obstacle density, as measured by the mean value of the FIFO queue of recent depth map predictions, and distance to target.}
\label{fig:meter_correlations}
\end{figure}

Fig.~\ref{fig:meter_correlations} indicates that CADENCE is predicting larger slimming factors when there is a higher density of nearby objects -- which further warrants that higher fidelity perception is needed to navigate through more complex scenarios. Conversely, smaller values of $\rho$ are predicted when the drone is closer to the target. This is likely because detailed path planning becomes less important as the target is more directly visible. There are several other environmental and navigation factors correlated to the predicted values of $\rho$. When collision avoidance is triggered, the adaptation model tends to predict a value of $\rho=0$ more than when avoidance is not triggered. This is possibly because the policy prioritizes moving around the object in front of it, and there is previous depth maps in the FIFO queue that can be used to navigate around it. When the navigation component is predicting large steps, corresponding to moving down a large open corridor (\textit{i.e.}, a road or between trees), it tends to also predict a smaller value of $\rho$. The predicted value of $\rho$ often matches the one from the previous step, implying that the model bases its predictions on persistent environmental conditions. The relationship between optimal values of $\rho$, navigation features, and environmental factors is quite complex. However, these correlations demonstrate that the adaptation logic is driven by learned context features rather than stochastic ones, warranting that the policy demonstrates a robust and interpretable adaptation strategy.

\section{Conclusions}
\label{sec:conclusion}

We presented CADENCE, an adaptive perception-and-control system designed to overcome the computational bottleneck of monocular depth estimation (MDE) on resource-constrained autonomous vehicles. By optimizing the trade-off between efficiency and reliability through utilizing a slimmable MDE network and a unified navigation-and-adaptation policy, CADENCE successfully closes the loop between sensing and actuation. Our approach moves beyond static model reduction by intelligently scaling computations in real-time, reserving high-fidelity perception for mission-critical navigation and environmental contexts.

Experimental results on a hardware-in-the-loop (HIL) testbed, using an NVIDIA Jetson Orin Nano with Microsoft AirSim, demonstrate that CADENCE achieves a 75.0\% reduction in energy expenditure, 16.1\% improvement in power consumption, and a 74.8\% decrease in inference latency compared to static state-of-the-art (SoTA) baselines. Crucially, these efficiency gains are accompanied by a 7.43\% improvement in navigation accuracy, proving that context-aware perception can actually improve reliability in autonomous flight, whereas model reduction approaches are typically accompanied with a degradation in accuracy. These findings highlight the potential for green IoT frameworks to enable autonomy on lightweight vehicles. This has implications in not only reducing mission failures due to depleted energy reserves or violated time constraints, but also improves flight time by reducing the total energy expenditure, reaction time to the environment by reducing the time spent computing, and power consumption which enables the device to concurrently host other operations.

\section*{ACKNOWLEDGMENTS}

This work was partially supported by the U. S. National
Science Foundation under Grant No. CCF-2140154.

\bibliographystyle{IEEEtran}
\bibliography{bibliography}

\end{document}